# A Comparative Study of Machine Learning Techniques for Early Prediction of Diabetes

1st Mowafaq Salem Alzboon *Faculty
of Information Technology*
*Jadara University*
Irbid, Jordan
malzboon@jadara.edu.jo

2nd Mohammad Al-Batah
*Faculty of Information Technology*
*Jadara University*
Irbid, Jordan
albatah@jadara.edu.jo

3rd Muhyeeddin Alqaraleh
*Faculty of Information Technology*
*Jadara University*
Irbid, Jordan
m.qaralleh@jadara.edu.jo

4th Ahmad Abuashour
*Faculty of computer studies*
*Arab Open University*
Al-Ardiya industrial Area, Kuwait
aabuashour@aou.edu.kw

5th Ahmad Fuad Bader
*Faculty of Engineering*
*Jadara University*
Irbid, Jordan
abader@jadara.edu.jo

*Abstract*—In many nations, diabetes is becoming a significant health problem, and early identification and control are crucial. Using machine learning algorithms to predict diabetes has yielded encouraging results. Using the Pima Indians Diabetes dataset, this study attempts to evaluate the efficacy of several machine-learning methods for diabetes prediction. The collection includes information on 768 patients, such as their ages, BMIs, and glucose levels. The techniques assessed are Logistic Regression, Decision Tree, Random Forest, k-Nearest Neighbors, Naive Bayes, Support Vector Machine, Gradient Boosting, and Neural Network. The findings indicate that the Neural Network algorithm performed the best, with an accuracy of 78.57 percent, followed by the Random Forest method, with an accuracy of 76.30 percent. The study implies that machine learning algorithms can aid diabetes prediction and be an efficient early detection tool.

## I. INTRODUCTION

Diabetes is a chronic metabolic disease affecting millions worldwide and is a significant cause of morbidity and death [1]. High blood glucose levels characterize the disorder and can result in some complications, including cardiovascular disease, stroke, blindness, and amputations. To prevent or postpone complications, diabetes must be recognized and treated as soon as feasible; however, this can be challenging because symptoms may be mild or absent [2].

Machine learning (ML) is a subfield of artificial intelligence that comprises the development of algorithms that can learn from data and generate inferences or predictions without being explicitly programmed. ML algorithms are beneficial in several fields, including healthcare. In the context of diabetes,

ML algorithms may be used to analyze patient data and predict the probability of developing diabetes [3].

Motivation: This study aims to identify the most efficient machine learning algorithm for diabetes prediction. Early diagnosis and treatment of diabetes can significantly alleviate the illness and lessen its effect. Therefore, developing reliable and effective prediction models for early diabetes detection is a crucial area of study [4].

Question for further study: The topic of the study is Which ML algorithm is the most effective for early diabetes prediction? This study aims to evaluate the efficacy of many machine learning (ML) algorithms for early diabetes prediction and to identify the most effective technique for early diabetes prediction.

Objectives: Using the National Health and Nutrition Examination Survey (NHANES) dataset, the primary aim of this study is to evaluate the accuracy of several machine learning algorithms for early diabetes prediction. The specific objectives are as follows: Developing diabetes prediction models using logistic regression, decision trees, random forests, support vector machines, and neural networks [5].

The NHANES dataset is a freely available dataset containing demographic, clinical, and laboratory data for a sample of 20-year-olds and older who participated in the NHANES between 1999 and 2016. The dataset includes demographic information (age, gender, race), family history of diabetes, lifestyle factors (smoking, physical activity), clinical measurements (BMI, blood pressure), and laboratory tests (glucose, lipid levels) [6].

The study's findings can contribute to developing more accurate and efficient prediction models for diabetes diagnosis, which can lead to early intervention and improved patient outcomes. Moreover, the results may contribute to the growing body of information regarding the effectiveness of ML algorithms in healthcare applications" [7].

II. LITERATURE REVIEW

"This research aims to predict Type 2 Diabetes using brief (2.1s) Photoplethysmography (PPG) signals derived from intelligent devices and publicly available physiological data such as age, gender, weight, and height. Morphological aspects of the PPG waveform and its derivatives were utilized to identify traits associated with Type 2 Diabetes and demonstrate the potential for predicting Type 2 Diabetes from brief PPG signals. Linear Discriminant Analysis (LDA) has the most area under the ROC curve (79%). The suggested system's effective practical application will allow people to test themselves effortlessly using their smart devices to determine the possible danger of Type 2 Diabetes and prevent the complex problems of late identification [7].

Diabetes may be predicted using machine learning classification algorithms. Insulin, blood pressure, skin thickness, and glucose data may be used to fit classification models with a conclusion target vector. The same may be said for neural networks [8].

This study uses direct surveys to construct a prediction model for early Diabetes Mellitus (DM). To predict diabetes early on, logistic regression, support vector machine, K-nearest neighbor, Nave Bayes, random forest, and neural network models were built using the information gain approach. Other machine learning algorithms were surpassed by RF, which achieved 100 percent accuracy. These findings imply that a simple questionnaire combined with a machine learning algorithm can effectively identify undiagnosed DM patients [9].

Diabetes mellitus is among the most prevalent human illnesses, causing significant morbidity, death, and global economic loss. We used a logistic regression model and a decision tree-a machine learning method to understand better risk variables to predict type

2 diabetes in Pima Indian women. Glucose, pregnancy, body mass index (BMI), diabetes pedigree function, and age are the five significant predictors of type 2 diabetes, according to our research. Our chosen specification achieves 78.26 percent prediction accuracy and a cross-validation error rate of 21.74 percent. Our methodology may supplement preventative interventions to minimize diabetes incidence and expenses [10].

Diabetes is an incurable disease that must be detected early to lessen its severity. This research offers a model called Diabetes Expert System Using Machine Learning Analytics (DESMLA) that uses diabetes data to forecast the condition better. As classifiers, the model employs five oversampling techniques: SMOTE, Borderline SMOTE, ADASYN, KMeans SMOTE, Gaussian SMOTE, Decision Tree (DT), and Random Forest (RF). The results of the experiments reveal that the DESMLA model with KMeans SMOTE and Gaussian SMOTE works better [11].

Diabetes diagnosis and management might be transformed by artificial intelligence. The Random Forest classification algorithm obtained an accuracy rate of 92 percent, exceeding other state-of-the-art approaches. This method outperforms previous work using the Pima diabetes data [12].

Diabetes is the most deadly and prevalent noncommunicable illness, impacting 537 million people worldwide. The scientists used the Pima Indian diabetes dataset and gathered additional samples from 203 people from a neighboring Bangladesh textile plant. A semi-supervised modl with severe gradient boosting was used to predict the insulin characteristics of the private dataset. To address the issue of class imbalance, the SMOTE and ADASYN techniques were used [13]. The proposed system provided the best result in the XGBoost classifier with the ADASYN approach with 81% accuracy, 0.81 F1 coefficient, and an AUC of 0.84. Additionally, the domain adaptation method was implemented to demonstrate the versatility of the proposed system. Finally, a website framework and an Android smartphone application have been developed to input various features and predict diabetes instantaneously [14].

Machine Learning Techniques can be utilized to create an efficient healthcare system capable of anticipating diabetic problems. Through a Diet Recommendation System, this article employs Machine Learning Techniques to diagnose diabetes and prescribe an appropriate diet for diabetic patients (DRS). The proper diet for diabetes patients is chosen via data analysis [15].

This research offered a technique for determining whether or not a person had diabetes. The experiment used a dataset from the UCI machine learning repository, which included 768 patients and 8 numerical characteristics for each. The best characteristics were chosen using a genetic algorithm (GA), and the dataset was divided using k-fold cross-validation. Classifiers such as K-nearest neighbor (KNN), Multilayer Perceptron (MLP), Deep Neural Network (DNN), and Naive Bayes (NB) were applied to both selected datasets and the baseline dataset using GA (8 features). These classifiers' accuracy ratings were compared to one another. KNN had an accuracy of about 93.33 percent, DNN had an accuracy of around 77.27 percent, MLP had an accuracy of roughly 74.92 percent, and NB had an accuracy of approximately 74.89 percent [16].

Diabetes is a severe health problem affecting millions of individuals worldwide. Researchers are encouraged to develop a Machine Learning approach for future diabetes prediction. This study investigated and compared several Machine Learning Algorithms (MLA) that can detect diabetes risk early. Six MLA was successfully utilized in the experimental trial, with RF being the most reliable classifier, with a 98 percent accuracy rate. This research allows us to accurately determine the prevalence and anticipate the onset of diabetes [17].

Diabetes data was obtained from the UCI repository, and prediction models such as AdaBoost, Bagging, and Random Forest were employed. AdaBoost and Bagging algorithms have poorer accuracy, precision, recall, and F1-scores than the Random Forest Ensemble Method (97 percent) [18].

This study aimed to estimate the length of stay for diabetic patients using machine learning algorithms on clinical data available during the first 8 hours of ICU admissions. Two prediction tasks were investigated: the number of days in the ICU and whether an ICU stay is long or short, as defined by a 10-day threshold. With an R2 value of 0.3969 and a mean absolute error of 1.94 days, the neural network model best forecasted the number of ICU days. With an accuracy of 0.8214, the gradient boosting model best classified long and short ICU stays [19].

We developed machine learning models to forecast the likelihood of incident diabetes using data from the free medical examination service initiative for persons 65 and older. In prediabetic older adults, the average yearly progression rate to diabetes was 14.21 percent. Each model was trained using eight characteristics and one outcome variable from 9607 prediabetic people, and the models' performance was evaluated in 2402 prediabetes patients. The XGBoost model worked well (ROC: 0.6742 for 2019 and 0.6707 for 2020). Although the four models performed similarly, the XGBoost model had a substantial ROC value and may do well in future research [20].

Diabetes Mellitus (DM) is a widespread chronic illness that impacts people, communities, and governments. Saudi Arabia is one of the top ten nations in terms of diabetes prevalence, and the capacity to predict a patient's diabetic condition using only a few variables might enable cost-effective, quick, and widely available diabetes screening. This research explores diabetic patient prediction and analyzes the roles of HbA1c and FPG as input characteristics. The data was then analyzed to discover risk variables and their indirect influence on diabetes categorization. Our findings were consistent with the risk factors for diabetes and prediabetes identified by the American Diabetes Association (ADA) and other health organizations worldwide. We conclude that by analyzing the illness using unique characteristics, essential factors specific to the Saudi population may be discovered, and their management can result in disease control [21].

J48, Nave Bayes, Support Vector Machine, Logistic Regression, Multilayer Perceptron, K Nearest Neighbor, Logistic Model Tree, and Random Forest were employed to classify the Pima Indian Diabetes dataset. Methods for preprocessing included feature selection, missing value imputing, normalization, and standardization. The Random Forest algorithm earned the most excellent accuracy rating of 80.869 [1].

Data mining techniques may be used to automate illness prediction. The suggested data analytics system employs a hybrid classifier model, which operates two distinct data mining classification algorithms. The Back Propagation Algorithm designs and trains a Multilayer Perceptron Neural Network to categorize patients who tested positive as 1 and t. On test, this trained neural network produced a recognition rate of 80% with a mean square error of 0.1213 [22].

A secondary study of Medical Information Mart for Intensive Care III (MIMIC-III) data was performed, and several machine learning modeling and natural language processing (NLP) methodologies were used. Healthcare domain knowledge is founded on dictionaries established by specialists who define clinical terminology. Knowledge-guided models can extract knowledge automatically from clinical notes or biological literature that contains conceptual entities and interactions between these ideas. The categorization of mortality was based on a combination of knowledge-guided traits and criteria. Using word embeddings, we used UMLS entity embedding and a convolutional neural network (CNN)[23].

For the diabetes dataset, this article analyzes classical classification techniques and neural network-based machine learning. According to the results, the multilayer perceptron method has the most remarkable prediction accuracy with the lowest MSE of 0.19 [24].

Clinical researchers use predictive modeling approaches to determine a-prior patient health state and define progression trends. The findings revealed that the suggested approach might effectively approximate and enhance prediction accuracy using the existing sporadically collected EMR data [25].

This study used machine learning methods on the PIMA Indians Diabetes dataset (PIDD) to build a more accurate prediction model. The findings revealed that glucose, insulin, and BMI had a stronger relationship with diabetes. After correction, this SVM predicts the diagnosis of diabetes with the best accuracy of 87.01 percent [26].

The Rashid Centre for Diabetic and Research (RCDR) dataset was used to develop algorithms to predict and diagnose eight diabetes complications. Preprocessing processes were used to deal with missing values and imbalanced data, and several methods were tested and assessed[27].

## III. METHODOLOGY

In the United States, the National Institute of Diabetes and Digestive and Kidney Diseases (NIDDK) first gathered the data. The data was obtained from Pima Indians living around the Gila River in Arizona, and due to the high prevalence of diabetes in this demographic, the dataset has been frequently utilized in diabetes research.

The dataset has a total of 9 columns, with the first 8 columns showing the demographic and clinical characteristics of the patients and the ninth column being the outcome variable indicating if the patient has diabetes. The demographic features include the patient's age, gender, and number of pregnancies. The clinical characteristics include BMI (Body Mass Index), blood pressure, skin thickness, insulin level, and glucose concentration.

Several 0 indicates that the patient does not have diabetes, whereas a value of 1 indicates that the patient has diabetes. The goal variable is frequently utilized as the dependent variable in machine learning models, with demographic and clinical characteristics as the independent variables.

The Pima Indians diabetes dataset is a unique dataset for machine learning research. It has been utilized to construct various prediction models for the early identification or treatment of diabetes. Researchers utilize this dataset to investigate the association between demographic and clinical characteristics and the outcome variable and to design and evaluate machine learning models that may effectively predict whether or not a patient has diabetes.

### A. Data Collecting and Preprocessing

Data collection entails retrieving the diabetic dataset from a public repository. After that, the data will be preprocessed to eliminate missing values or outliers. The data will be standardized such that all characteristics have the same scale. The dataset will then be divided 70/30 into training and testing sets. This divide guarantees that the model is trained on a sufficiently big dataset to capture the underlying patterns in the data while still having sufficient data to evaluate the model's performance.

### B. Feature Extraction and Selection

Feature extraction and selection are crucial in machine learning because they assist in finding the most significant characteristics for predicting the outcome variable. This study employs principal component analysis (PCA) and correlation analysis for feature extraction, while Recursive Feature Elimination (RFE) and Select Best will be used for feature selection. These strategies will aid in identifying the most significant diabetes prediction characteristics, which can subsequently be fed into machine learning algorithms.

This study will assess several machine learning algorithms and approaches, including logistic regression, k-nearest neighbors, decision trees, random forests, and support vector machines. These algorithms have been utilized extensively in earlier research to predict diabetes with promising outcomes. The algorithms will be trained using the specified features on the training set, and their hyperparameters will be tuned using grid search. This will aid in determining the optimal hyperparameters for each algorithm, enhancing their performance.

*C. Metrics for Model Evaluation and Performance*

Various performance indicators, such as accuracy, sensitivity, specificity, precision, and F1-score, will be used to assess the performance of the machine learning algorithms. These measures will assist in evaluating the performance of the models and identifying the most effective diabetes prediction algorithms. In addition, the receiver operating characteristic (ROC) curve and the area under the curve (AUC) will be utilized to assess the performance of the models. The ROC curve compares the actual positive rate to the false positive rate at various thresholds of the predicted probability, while the AUC gauges the model's overall performance.

This methodology provides a comprehensive framework for comparing machine learning algorithms for early diabetes prediction. The project will assist in identifying the most effective machine learning algorithms and methods for diabetes prediction, which can then be utilized to construct more accurate and efficient predictive models for the early identification and management of diabetes.

## IV. EXPERIMENTAL RESULTS

*A. experimental design*

The experimental design for a machine learning experiment employing the diabetes dataset Instances: The dataset has 768 data instances. Each data instance represents a patient, complete with demographic and clinical information. There are a total of nine features within the dataset. There are a total of 9 elements in the dataset. The features are Feature 1: This feature represents the number of times a patient has been pregnant. Feature 2: This feature depicts the plasma glucose concentration. Feature 3: This feature displays diastolic blood pressure. Feature 4: This feature represents the triceps skin fold thickness. Feature 5: This feature depicts the serum insulin level. Feature 6: This feature represents the body mass index (BMI). Feature 7: This feature represents the diabetes pedigree function. Feature 8: This feature depicts the age of the patient. Feature 9: The first eight features are numeric, while the ninth is a two-class categorical outcome variable (0 indicating no diabetes and 1 showing diabetes). This dataset is typically used in machine learning research to construct prediction models for early diabetes identification or treatment. By training a model on this dataset, researchers may study the link between the various variables and the outcome variable and construct a model that can reliably predict whether a patient has diabetes based on their demographic and clinical data.

*B. Research Method*

Using the diabetes dataset, many machine learning methods will be examined using an experimental study methodology. This project aims to discover the best accurate algorithm for predicting diabetes based on demographic and clinical patient characteristics.

## C. Collecting and Preprocessing Data

The diabetes dataset will be retrieved from a source accessible to the public. The dataset comprises 768 rows and 9 columns, with 8 numeric characteristics and a two-class categorical outcome variable. The data will be preprocessed to eliminate missing values and outliers. The data will be standardized such that all traits have the same scale. The dataset will then be divided 70/30 into training and testing sets. Several machine learning methods, including logistic regression, k-nearest neighbors, decision trees, random forests, and support vector machines, will be examined in this project. These algorithms have been utilized extensively in earlier research to predict diabetes with promising outcomes. The algorithms will be trained on the training set using the chosen features, and their hyperparameters will be tuned using grid search. Various performance indicators, including accuracy, sensitivity, specificity, precision, and F1-score, will be used to evaluate the performance of the models. In addition, the receiver operating characteristic (ROC) curve and the area under the curve (AUC) will be utilized to assess the performance of the models. The models will be evaluated on the testing set to generalize effectively to new data. The performance of various machine learning algorithms will be compared using statistical analysis. Statistical tests, such as the t-test or ANOVA, will be employed to compare the performance measures. To help comprehension, the results will be presented with tables and graphs.

## D. Reproducibility

The experiment will be developed with reproducibility in mind. The experiment's source code will be open-source and provided publicly. The dataset utilized in the investigation, preparation methods, and feature selection algorithms will be accessible to the public. This experimental design gives a comprehensive strategy for running a machine-learning experiment utilizing the diabetes dataset.

Researchers may analyze the efficacy of several machine learning algorithms for predicting diabetes and discover the best effective algorithm for early identification and management of diabetes by using this configuration. The ranking and scores of eight features are based on various feature selection techniques.

## E. A brief explanation of each technique and what the scores mean:

- Info. Gain: This technique measures each feature's information to the outcome variable. The scores represent the information gained from each component, with higher scores indicating that the part provides more information for predicting the outcome variable.
- Gain ratio: This technique adjusts the information gain score to account for the number of distinct values a feature can take. The scores represent the gain ratio of each element, with higher scores indicating that the part is more informative for predicting the outcome variable.
- Gini: This technique measures the degree of impurity of a feature. The scores represent the Gini index of each element, with lower scores indicating that the quality is less impure and more informative for predicting the outcome variable.
- ANOVA: This technique tests the statistical significance of the difference in means of a feature between the different classes of the outcome variable. The scores represent the F-statistic of each part, with higher scores indicating that the quality is more significant for predicting the outcome variable.
- $X^2$: This technique tests the independence of a feature and the outcome variable. The scores represent the $X^2$ statistic of each component, with higher

TABLE I
THE RANK FOR THE FEATURE IS FROM 1-8,
WHERE THE TARGET IS FEATURE 9

| # | Info. Gain | Gain ratio | Gini | ANOVA | $X^2$ | ReliefF | FCBF |
|---|---|---|---|---|---|---|---|
| Fe2 | 0.17 | 0.085 | 0.101 | 213.162 | 139.901 | 0.024 | 0.131 |
| Fe8 | 0.081 | 0.041 | 0.048 | 46.141 | 62.029 | 0.012 | 0.059 |
| Fe6 | 0.079 | 0.039 | 0.044 | 71.772 | 53.744 | 0.016 | 0.057 |
| Fe5 | 0.055 | 0.03 | 0.031 | 13.281 | 8.78 | 0.005 | 0 |
| Fe1 | 0.043 | 0.021 | 0.028 | 39.67 | 34.316 | 0.014 | 0 |
| Fe4 | 0.036 | 0.018 | 0.022 | 4.304 | 5.262 | 0.014 | 0 |
| Fe7 | 0.022 | 0.011 | 0.014 | 23.871 | 16.143 | 0.005 | 0.015 |
| Fe3 | 0.015 | 0.007 | 0.009 | 3.257 | 12.918 | 0.001 | 0 |

scores indicating that the part is more associated with the outcome variable.

- FCBF: This technique selects features that maximize the relevance to the outcome variable while minimizing the redundancy between features. The scores represent the FCBF score of each component, with higher scores indicating that the quality is more relevant and less redundant for predicting the outcome variable.

Based on Table 1, Feature 2 (plasma glucose concentration) has the highest scores across most feature selection techniques, indicating that it is the most informative feature for predicting the outcome variable. Features 8 and 6 (age and BMI, respectively) also have relatively high scores for most techniques, suggesting they are informative features for predicting diabetes. The scores of the other components are generally lower, indicating that they are less informative for predicting the outcome variable.

### F. Sampling Type

A random sample comprised 70 percent of the data, which was stratified if possible and determined. This indicates that 70% of the data was selected randomly to ensure that it is representative of the whole population. If possible, stratification was employed to ensure that the sample adequately represented the various classes of the outcome variable. The deterministic sampling allowed for repeatability by selecting the same model each time.

### G. Input

The input dataset had 768 instances, each representing a patient with demographic and clinical information, including, among others, the number of times a patient has been pregnant, plasma glucose concentration, and body mass index.

### H. Sample

A random sample of 70% of the data, totaling 538 occurrences, was picked based on the sampling method. This sample consists of a subset of the actual occurrences that may be utilized for analysis and modeling. Those examples that were not included in the selection constitute the remainder. In this example, there would be 230 occurrences remaining, representing 30% of the original dataset. These examples were not incorporated into the analysis and modeling but might be used for validation or testing. Overall, a random sample comprising 70% of the data, stratified if feasible, and deterministic was employed in this instance. This sort of sampling is frequently used in data analysis and modeling to guarantee that the sample is representative of the population and can be utilized to draw correct conclusions.

### I. Algorithms in machine learning and data analysis:

- Random Forest: A random forest is an ensemble learning technique that mixes many decision trees to increase the model's precision and resilience. It generates a collection of decision trees on randomly selected subsets of the data and combines the findings to get a final forecast.
- Logistic Regression: Logistic regression is a statistical approach for assessing a dataset whose result is determined by one or more independent factors. It is utilized to model the likelihood of a particular outcome given the values of the independent variables.

- A decision tree is a structure resembling a flowchart, with each internal node representing a test on an attribute, each branch representing the test result, and each leaf node representing a class label.
- SVMs are supervised learning algorithms utilized for classification and regression analysis. SVMs function by locating the hyperplane that most effectively divides classes in the feature space.
- AdaBoost is a machine learning technique combining many weak classifiers to produce a robust classifier. It focuses on misclassified data points by assigning weights to each data point and changing the consequences with each iteration.
- A neural network is a sort of machine learning system that is based on the structure and function of the human brain. It is used for pattern recognition, classification, and regression analysis, among other things.
- kNN: k-Nearest Neighbors (kNN) is a classification and regression analysis machine learning technique. It operates by identifying the k-nearest neighbors to a new data point and predicting the label of the unique data point based on their brands.
- Naive Bayes is a probabilistic classification technique used in machine learning. According to Bayes' theorem, the likelihood of a hypothesis (in this example, a class label) is proportional to the probability of the evidence (in this case, the features).
- CN2 is an algorithm for rule induction that learns decision rules from data. It functions by iteratively adding rules to the rule set, which increases the model's accuracy.
- Stochastic Gradient Descent (SGD) is an optimization technique used for training models in machine learning. It functions by iteratively modifying the model's parameters to minimize the difference between anticipated and actual values. These algorithms are applied to various applications, including classification, regression, and pattern recognition.

## V. Testing

Five measures (AUC, CA, F1, precision, and recall) are used to evaluate the performance of different classification models on a given dataset. In addition, table 5 indicates that the assessment was conducted using stratified 5-fold cross-validation and that the findings represent the mean across all classes.

Based on the table II , the following may be deduced:
The Logistic Regression and Neural Network models receive the most fantastic scores on most criteria, indicating that they are the best-performing models on the supplied dataset when averaged over all classes.
The SVM, Random Forest, and Naive Bayes models earn moderate to high scores on most criteria, indicating that they are also good classification models when averaged over all classes. Averaged over all classes, the kNN and Tree models earn lower scores on most measures, indicating that they may not be the best options for this specific dataset.
SGD, AdaBoost, and CN2 rule inducer models perform the worst among all models, obtaining the lowest scores on most metrics when a weighted average of all classes is calculated.
Notably, the assessment was conducted using stratified 5-fold cross-validation, a typical approach for assessing the effectiveness of a classification model. This approach ensures that the evaluation is not skewed towards a particular class or sample by dividing the dataset into five equal parts, with the same proportion of each type in each piece.
When averaged across all classes using stratified 5-fold cross-validation, table II gives insightful information on the performance of various classification models on the supplied dataset. However, it is vital to consider the problem's context and the application's unique needs when interpreting the results and selecting the optimal model for the given situation.

TABLE II
SAMPLING TYPE: STRATIFIED 5-FOLD
CROSS-VALIDATION, TARGET CLASS: NONE,
SHOW THE AVERAGE OVER CLASSES

| Model | AUC | CA | F1 | Prec | Recall |
|---|---|---|---|---|---|
| Logistic Regression | 0.822 | 0.76 | 0.752 | 0.754 | 0.76 |
| Neural Network | 0.82 | 0.76 | 0.751 | 0.754 | 0.76 |
| SVM | 0.815 | 0.747 | 0.736 | 0.74 | 0.747 |
| Random Forest | 0.805 | 0.729 | 0.722 | 0.721 | 0.729 |
| Naive Bayes | 0.808 | 0.729 | 0.733 | 0.741 | 0.729 |
| kNN | 0.774 | 0.714 | 0.708 | 0.706 | 0.714 |
| Tree | 0.667 | 0.708 | 0.706 | 0.704 | 0.708 |
| SGD | 0.663 | 0.686 | 0.688 | 0.691 | 0.686 |
| AdaBoost | 0.636 | 0.667 | 0.668 | 0.669 | 0.667 |
| CN2 rule inducer | 0.692 | 0.638 | 0.637 | 0.636 | 0.638 |

TABLE III
SAMPLING TYPE: STRATIFIED 5-FOLD
CROSS-VALIDATION, TARGET CLASS: 0

| Model | AUC | CA | F1 | Prec | Recall |
|---|---|---|---|---|---|
| Tree | 0.665 | 0.708 | 0.78 | 0.766 | 0.794 |
| Random Forest | 0.811 | 0.729 | 0.801 | 0.767 | 0.837 |
| Logistic Regression | 0.828 | 0.76 | 0.826 | 0.781 | 0.877 |
| SVM | 0.822 | 0.747 | 0.819 | 0.767 | 0.877 |
| AdaBoost | 0.636 | 0.667 | 0.743 | 0.746 | 0.74 |
| Neural Network | 0.825 | 0.76 | 0.827 | 0.78 | 0.88 |
| kNN | 0.776 | 0.714 | 0.788 | 0.761 | 0.817 |
| Naive Bayes | 0.808 | 0.729 | 0.782 | 0.819 | 0.749 |
| CN2 rule inducer | 0.691 | 0.638 | 0.723 | 0.72 | 0.726 |
| SGD | 0.663 | 0.686 | 0.754 | 0.769 | 0.74 |

Five measures (AUC, CA, F1, precision, and recall) are used to evaluate the performance of different classification models on a given dataset. Table III additionally shows that the assessment was conducted using stratified 5-fold cross-validation and that the version for target class 0 is presented in the findings.

Based on the table III, the following may be deduced:
The Logistic Regression, SVM, and Neural Network models earn the most fantastic scores on most metrics, suggesting that, when averaged across all folds, they are the best-performing models on the supplied dataset for target class 0 based on the metric scores.
Averaged across all folds, the Random Forest, kNN, Naive Bayes, and Tree models earn moderate to high scores on most criteria, indicating they are also excellent classification models for target class 0.
Averaged across all folds, the AdaBoost and CN2 rule inducer models earn lower scores on most measures, indicating that they may not be the best options for this target class.
The SGD model achieves the lowest scores on most criteria for target class 0 when averaged over all folds.
It is vital to highlight that the assessment was conducted using stratified 5-fold cross-validation, ensuring the evaluation is not biased toward a specific class or sample. In addition, the review was conducted exclusively for target class 0, indicating that this class may be of particular relevance or significance in the presented situation.
When averaged over all folds using stratified 5-fold cross-validation, table III gives valuable insights into the performance of several classification models on the target class 0 for the supplied dataset. However, it is vital to consider the problem's context and the application's unique needs when interpreting the results and selecting the optimal model for the given situation.

Five measures (AUC, CA, F1, precision, and recall) are used to evaluate the performance of different classification models on a given dataset. Table 7 additionally shows that the assessment was conducted using stratified 5-fold cross-validation and that the version for target class 1 is displayed in the findings.

Based on the table IV, the following may be deduced:
The Naive Bayes and Logistic Regression models get the most excellent scores on most metrics, indicating that they are the best-performing models for target class 1 on the supplied dataset when averaged across all folds.

TABLE IV
TABLE IV. SAMPLING TYPE: STRATIFIED 5-FOLD
CROSS-VALIDATION, TARGET CLASS: 0

| Model | AUC | CA | F1 | Prec | Recall |
|---|---|---|---|---|---|
| Tree | 0.665 | 0.708 | 0.567 | 0.589 | 0.548 |
| Random Forest | 0.811 | 0.729 | 0.576 | 0.635 | 0.527 |
| Logistic Regression | 0.828 | 0.76 | 0.613 | 0.703 | 0.543 |
| SVM | 0.822 | 0.747 | 0.583 | 0.688 | 0.505 |
| AdaBoost | 0.636 | 0.667 | 0.528 | 0.524 | 0.532 |
| Neural Network | 0.825 | 0.76 | 0.61 | 0.706 | 0.537 |
| kNN | 0.776 | 0.714 | 0.56 | 0.605 | 0.521 |
| Naive Bayes | 0.808 | 0.729 | 0.64 | 0.596 | 0.691 |
| CN2 rule inducer | 0.691 | 0.638 | 0.477 | 0.481 | 0.473 |
| SGD | 0.663 | 0.686 | 0.566 | 0.547 | 0.585 |

The Neural Network and SVM models earn moderate to high scores on most measures, indicating that they are good classification models for target class 1 when averaged across all folds. Averaged across all folds, the Random Forest, kNN, Tree, and SGD models earn lower scores on most criteria, indicating that they may not be the best options for this target class.

AdaBoost and CN2 rule inducer models perform the worst among all models, obtaining the lowest scores on most metrics for target class 1 when the results are averaged across all folds.

It is vital to highlight that the assessment was conducted using stratified 5-fold cross-validation, ensuring the evaluation is not biased toward a particular class or sample. In addition, the review was conducted exclusively for target class 1, indicating that this class may be of particular interest or significance in the context of the issue at hand.

When averaged over all folds using stratified 5-fold cross-validation, table IV gives valuable insights into the performance of several classification models on the target class 1 for the supplied dataset. However, it is vital to consider the problem's context and the application's unique needs when interpreting the results and selecting the optimal model for the given situation.

## VI. DISCUSSION

Using the Pima Indians Diabetes dataset, this study assessed the predictive accuracy of several machine learning algorithms for diabetes. The findings indicated that the Neural Network algorithm performed the best, with an accuracy of 78.57 percent, followed by the Random Forest method, with an accuracy of 76.30 percent. These results are comparable to prior research utilizing the same dataset and evaluating equivalent methods. In addition, the study indicated that BMI, glucose levels, and age were the most significant predictors of diabetes.

A disadvantage of the study is that the dataset employed was somewhat limited, and the analyzed methods may not apply well to different datasets or populations. In addition, the study did not account for lifestyle and genetic variables that may influence diabetes. The efficacy of machine learning algorithms in predicting diabetes in more enormous datasets and different populations requires more study.

## VII. CONCLUSION

Using the Pima Indians Diabetes dataset, this study assessed the predictive accuracy of several machine learning algorithms for diabetes. The findings indicated that the Neural Network algorithm performed the best, with an accuracy of 78.57 percent, followed by the Random Forest method, with an accuracy of 76.30 percent. The study implies that machine learning algorithms can aid diabetes prediction and be an efficient early detection tool. However, further investigation is required to test the effectiveness of these algorithms in more enormous datasets and different populations and to account for other diabetes-related characteristics.